%% file: PaperForReview.tex
\documentclass[10pt,twocolumn,letterpaper]{article}

\usepackage[pagenumbers]{wacv} 

\usepackage[accsupp]{axessibility}
\usepackage{graphicx}
\usepackage{amsmath}
\usepackage{amssymb}
\usepackage{booktabs}
\usepackage{multirow}
\usepackage{enumitem}

\usepackage[pagebackref,breaklinks,colorlinks]{hyperref}

\DeclareMathOperator*{\argmin}{arg\,min}

\makeatletter
\newcommand*\bigcdot{\mathpalette\bigcdot@{2}}
\newcommand*\bigcdot@[2]{\mathbin{\vcenter{\hbox{\scalebox{#2}{$\m@th#1\bullet$}}}}}
\makeatother

\usepackage[capitalize]{cleveref}
\crefname{section}{Sec.}{Secs.}
\Crefname{section}{Section}{Sections}
\Crefname{table}{Table}{Tables}
\crefname{table}{Tab.}{Tabs.}

\def\wacvPaperID{1225} 
\def\confName{WACV}
\def\confYear{2025}

\begin{document}

\title{TaCOS: Task-Specific Camera Optimization with Simulation}

\author{Chengyang Yan \qquad Donald G. Dansereau\\
University of Sydney\\
{\tt\small {chengyang.yan, donald.dansereau}@sydney.edu.au}
}
\maketitle

\begin{abstract}
   The performance of perception tasks is heavily influenced by imaging systems. However, designing cameras with high task performance is costly, requiring extensive camera knowledge and experimentation with physical hardware. Additionally, cameras and perception tasks are mostly designed in isolation, whereas recent methods that jointly design cameras and tasks have shown improved performance. Therefore, we present a novel end-to-end optimization approach that co-designs cameras with specific vision tasks. This method combines derivative-free and gradient-based optimizers to support both continuous and discrete camera parameters within manufacturing constraints. We leverage recent computer graphics techniques and physical camera characteristics to simulate the cameras in virtual environments, making the design process cost-effective. We validate our simulations against physical cameras and provide a procedurally generated virtual environment. Our experiments demonstrate that our method designs cameras that outperform common off-the-shelf options, and more efficiently compared to the state-of-the-art approach, requiring only 2 minutes to design a camera on an example experiment compared with 67 minutes for the competing method. Designed to support the development of cameras under manufacturing constraints, multiple cameras, and unconventional cameras, we believe this approach can advance the fully automated design of cameras. Code is available on our project page at \href{https://roboticimaging.org/Projects/TaCOS/}{\textcolor{magenta}{https://roboticimaging.org/Projects/TaCOS/}}.
\end{abstract}

\section{Introduction}
\label{sec:intro}
The quality of camera captures directly affects perception tasks. An analogy in nature is how animals' visual perceptions impact their daily activities. It is believed that evolution designs distinct visual systems for different species to suit their habitats~\cite{land2012animal}, suggesting a need for a sophisticated and application-based camera design approach.

Currently, designing cameras is cumbersome, typically requiring professionals to devise designs based on their experiences, followed by extensive experiments with various design decisions. Moreover, the camera hardware is often designed in isolation from perception tasks, despite literature showing the benefits of joint design.

\begin{figure}
      \centering
      \includegraphics[width=\linewidth]{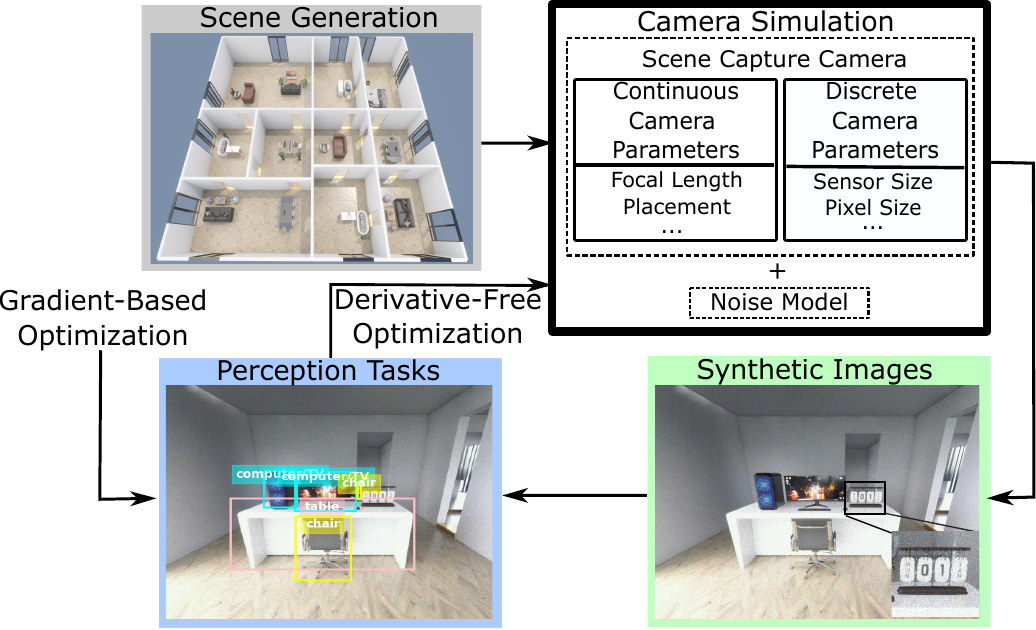}
      \caption{Our method combines a derivative-free optimizer and gradient-based optimizer to co-design the camera with perception tasks in simulation, which utilizes ray tracing and a physics-based noise model. Our approach supports optimising discrete and continuous camera parameters for manufacture constraints and the generalization to other camera design problems.}
      \label{fig:key_figure}
\end{figure}

Existing methods optimize imaging systems for tasks through software, often using ray tracing renderers to simulate the physical image formation process. Ray tracing render has been widely used for designing and evaluating cameras for autonomous driving~\cite{blasinski2018optimizing, liu2019system, liu2019soft, weikl2021end, liu2023using}. However, manual tuning of camera parameters for optimization is still required, and joint optimization of the cameras and the tasks is not addressed in these methods.

Other works propose propose automatic co-design of imaging systems and perception tasks using differentiable ray tracing or proxy neural networks~\cite{sitzmann2018end, sun2021end, yang2024curriculum, chang2019deep, ikoma2021depth, baek2021polka, tseng2019hyperparameter, tseng2021differentiable, robidoux2021end, cote2023differentiable, metzler2020deep, martel2020neural, chang2018hybrid, diamond2021dirty, zhang2022all, yang2023image, nguyen2022learning}. Nevertheless, these methods focus primarily on optics design since their design space is restricted by using captured image datasets which prevents them from generalizing to more complex camera design problems involving field-of-view (FOV), resolution, multi-cameras, or unconventional cameras.

Game engines like Unreal Engine (UE)~\cite{unrealengine} are repurposed for simulating cameras due to their ray tracing support and ability to produce video sequences and interact with virtual environments. Klinghoffer et al.~\cite{klinghoffer2023diser} propose a reinforcement learning (RL) method for camera design evaluated with a UE-based simulator. While this method achieves high performance, it optimizes a complex neural network to design cameras, making it less efficient than directly optimizing camera parameters.

As illustrated in Fig.~\ref{fig:key_figure}, we introduce an end-to-end camera design method that directly optimizes camera parameters for perception tasks. We propose a camera simulator that allows tuning various parameters and addresses image signal-to-noise ratio (SNR) with a physics-based noise model\cite{foi2009clipped}. Additionally, we use a procedural generation algorithm to create indoor virtual environments in UE 5 with machine learning labels for our method.

Inspired by evolution, we employ a genetic algorithm~\cite{holland1992adaptation}, a derivative-free optimizer, to optimize discrete, continuous, and categorical variables, which supports manufacturing constraints and selecting parts (optics, image sensors, etc.) from catalogs where custom manufacture is infeasible. Finally, we adopt a quantized continuous approach~\cite{cote2023differentiable} for discrete variables, considering their interdependencies.

We implement our approach on two design problems, demonstrating our method's efficiency compared to the state-of-the-art (SOTA) design method, and cameras designed by our approach achieve compelling performance compared to high-quality robotic/machine vision cameras. We also validate our simulation's accuracy by comparing it with physical cameras. In summary, our contributions are:
\begin{itemize}[noitemsep]
    \item We introduce an end-to-end camera design method that combines derivative-free and gradient-based optimization to automatically co-design cameras with perception tasks, allowing continuous, discrete, and categorical camera variables.
    \item We develop a camera simulation with a physics-based noise model and a virtual environment, and provide a procedurally generated virtual environment.
    \item We validate our simulator by establishing equivalence in both low-level image statistics and high-level task performance between synthetic and captured imagery.
    \item We validate our method with improved performance compared to other methods and off-the-shelf options.
\end{itemize}

This work is a key step in simplifying the process of designing cameras for autonomous systems like robots, emphasizing task performance and manufacturability constraints.

\smallskip \noindent \textbf{Limitations}\hspace{2pt} Our work focuses exclusively on camera parameters set during the manufacturing stage. Parameters that can be dynamically adjusted during camera operation, such as exposure settings, will be addressed in future work. Nevertheless, our method can still select algorithms that control these dynamic parameters dynamically.

\section{Related Work}
\label{sec:related}
\smallskip \noindent \textbf{Task-Specific Simulation-Based Camera Design}\hspace{2pt} Designing cameras tailored for vision tasks using simulation has gained popularity. Blasinski et al.~\cite{blasinski2018optimizing} propose optimizing a camera design to detect vehicles. They used synthetic data generated with ISET~\cite{Wandell2024ISET3D, Wandell2024ISETCam}. The method optimizes cameras by experimentally analyzing the impact of image postprocessing pipelines and auto-exposure algorithms on object detection tasks. This work continues in \cite{liu2019system, liu2019soft, weikl2021end, liu2023using} with larger datasets and an optimization framework for high dynamic range (HDR) imaging. Nevertheless, these methods require manual tuning and testing of camera parameters, whereas our method is an automatic end-to-end approach.

\begin{figure*}
        \centering
        \includegraphics[width=\linewidth]{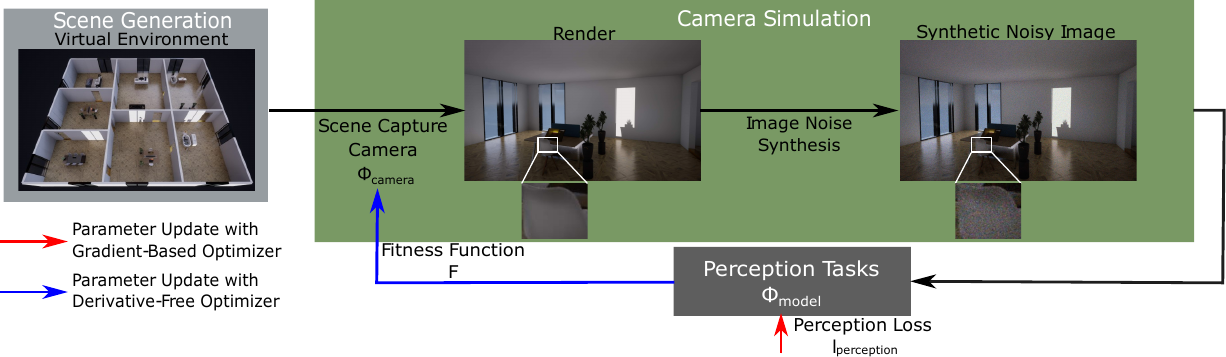}
        \caption{We establish a virtual environment and capture scene renders using a ray-traced scene capture camera. We then add physics-based, sensor-specific noise to the renders and input them into perception tasks for evaluation. In our optimization process, we jointly optimize the camera parameters $\Phi_{camera}$ using a fitness function $F$ with a derivative-free optimizer (blue arrow), as well as the parameters of perception tasks $\Phi_{model}$ (if trainable) on their corresponding loss function $l_{perception}$ with gradient-based optimizers (red arrow).}
\label{fig:method_overview}
\end{figure*}

Other works have proposed end-to-end methods to optimize the imaging system based on tasks. Many use gradient-based optimization with differentiable camera simulators, including differentiable ray tracing and proxy neural networks for non-differentiable image formation processes. In these works, prerecorded images are used as input scenes to their pipeline, and their simulators convert the input images to images formed by their proposed cameras. Applications include extended depth of field (DOF)~\cite{sitzmann2018end, sun2021end, yang2024curriculum}, depth estimation~\cite{he2018learning, chang2019deep, ikoma2021depth, baek2021polka}, object detection~\cite{tseng2019hyperparameter, tseng2021differentiable, robidoux2021end, cote2023differentiable}, HDR imaging~\cite{metzler2020deep}, \cite{martel2020neural}, image classification~\cite{chang2018hybrid, diamond2021dirty, zhang2022all, yang2023image}, and motion deblurring~\cite{nguyen2022learning}. However, these methods use precaptured images so that their camera simulation is restricted to the domains of the data. Key camera design decisions such as the FOV, resolution, use of multiple cameras, and the design of unconventional cameras (light field, etc.), are not addressed. Our method establishes a virtual environment to support the simulation and optimization of a much broader range of camera designs.

RL has also been explored for end-to-end optimization of imaging systems. Klinghoffer et al.~\cite{klinghoffer2023diser} uses RL to train a camera designer, encompassing various camera parameters using the CARLA Simulator~\cite{dosovitskiy2017carla}. Hou et al.~\cite{hou2023optimizing} introduce another RL-based approach for pedestrian detection. Although RL demonstrates impressive results in camera design, it involves optimizing complex neural networks that learn to design cameras, demanding more training data and time since the neural network contains a larger number of parameters that need to be optimized. In contrast, our approach directly optimizes the camera's parameters, yielding competitive results with much less computation.

\smallskip \noindent \textbf{Genetic Algorithms for Camera Design}\hspace{2pt} Genetic algorithms~\cite{holland1992adaptation} are widely applied across many domains thanks to their flexibility, robustness, and ability to explore complex search spaces including continuous, discrete, and categorical variables. For imaging systems design, existing works have applied genetic algorithms for optimizing camera placements in a network of cameras~\cite{olague2002optimal, aissaoui2018designing, heyns2021optimisation}, and for optimizing optics design to improve image quality~\cite{betensky1993postmodern, van1996optical, ono2000optimal, fang2007eliminating, fang2009extended, tsai2015improvement, yen2015aspherical}, a review can be found in \cite{hoschel2019genetic}. To our knowledge, none of these existing methods combines genetic algorithms with gradient-based optimizers to co-design the camera hardware and perception tasks for automatic camera design and improved performance, which is the primary contribution of our work.

\section{Method}
\label{sec:method}
Our method (Fig.~\ref{fig:method_overview}) uses a simulated camera to capture a virtual environment and apply a physics-based noise model. We then evaluate the resulting images in perception tasks and use this data to optimize the camera design. Our method outputs the designed camera parameters and trained models for perception tasks if the task is trainable.

\subsection{Simulation Environment}
\label{sec:simulation_env}
The simulation environment should be photorealistic to minimize the sim-to-real gap. It should support rendering with various camera parameters and illuminations to maximize the design space, as well as the deployment of multiple cameras for unconventional imaging systems.

Considering the desired features of the simulation environment, we chose to use UE for this work. UE utilizes real-time hardware ray tracing combined with software ray tracing for global illumination and reflection, aligning with the physics of light and providing realistic shadowing, ambient occlusion, illumination, reflections, etc.~\cite{unrealengine}. UE supports the alteration of various camera parameters, detailed in Sec.~\ref{sec:camera_model}, and the application of multiple cameras. Nevertheless, our method is not limited to UE, other simulators with the desired features or suited to the downstream tasks can also be used to build the simulation environment.

We deploy cameras on an auto-agent simulating the platform that uses the camera. The agent navigates the virtual environment autonomously, enabling a fully automated design process. The cameras capture scene renders as the agent moves, which are then used in downstream processes.

To demonstrate our method, we implemented the procedural generation of random indoor virtual environments and their associated semantic labels with UE 5, with support for application-specific objects. The implementation details of the environment are further explained in Sec.~\ref{sec:mono_mr}.

\subsection{Camera Simulation}
\label{sec:camera_model}
Our camera simulation includes a scene capture component from the simulation environment and an image noise synthesis component to augment the scene captures to enhance the realism of our simulation.

\smallskip \noindent \textbf{Scene Capture}\hspace{2pt} The scene capture component captures scene irradiance from the virtual environment. We employ the camera in UE that captures the scene renders, which allows the configuration of parameters associated with camera placement (location, orientation), optics (focal length and aperture), the image sensor (width, height, and pixel count), exposure settings (shutter speed and ISO), and multi-camera designs (number of cameras and their poses). It also allows the configuration of algorithms in the image processing pipeline (auto-exposure, white balancing, tone mapping, color correction, gamma correction, and compression) and includes image effects like motion blur.

We experimentally validate our method using parameters supported by UE. Additional parameters like geometric distortion and defocus blur could be added by augmenting the renderer. The noise synthesis model described below serves as an example of such augmentation.

\smallskip \noindent \textbf{Noise Synthesis}\hspace{2pt} Image noise is a fundamental limiting factor for many vision tasks that is tightly coupled to camera parameters such as pixel size and exposure settings. As UE lacks a realistic noise model, we incorporate a post-render image augmentation that introduces noise. We employ thermal and signal-dependent Poisson noise following the affine noise model~\cite{foi2009clipped}. The noise model is calibrated with a physical camera following established methods~\cite{ratner2007illumination, liu2006noise, wang2021multiplexed} and then generalized for different exposure time and gain settings. The generalized form of the noise model defines the variation of intensity at a pixel as
\begin{equation}\label{eqn:noise_generlization}
\sigma^2 = \frac{G}{G_0}\sigma_p^2 \Bar{I} + \frac{G^2}{G_0^2}\sigma_{r}^2,
\end{equation}
where $\sigma_p$ and $\sigma_r$ are photon and thermal noise respectively, $G$ and $G_0$ are the new gain and calibrated gain, and $\Bar{I}$ is the measured intensity. We detail the noise model calibration and generalization method in the supplementary material.

To generalize the noise model to different image sensors, we consider the ratio of their pixel sizes. For the same illumination, larger pixel sizes capture more photons, resulting in a higher measured intensity level, which is readily reflected by adjusting the gain in Eq.~\ref{eqn:noise_generlization} inversely proportional to the pixel area. While we employ these observations to generalise noise characterisations, it is also possible to directly characterize multiple sensors and directly use these characterizations for more accurate noise characteristics.

\subsection{Optimization}
\label{sec:joint_opt}
\smallskip \noindent \textbf{Optimizers}\hspace{2pt} We employ a derivative-free optimizer to optimize the camera parameters as many simulators and tasks used for designing cameras are non-differentiable, and numerous camera parameters are discrete or categorical, which makes gradient-based optimizers, such as gradient-descent, and gradient estimation based method, such as surrogate gradient and finite differences, inapplicable. Hence, we utilize the genetic algorithm~\cite{holland1992adaptation} as the derivative-free optimizer, although other derivative-free optimizers can also be integrated into the proposed pipeline.

To enhance the performance of perception tasks, we jointly optimize the tasks (if applicable) along with the design of the camera hardware. The applicable tasks are the ones that involve machine learning models and are optimized using their corresponding optimizers. Given that neural networks are commonly used in SOTA methods for these tasks, gradient-based optimizers such as Stochastic Gradient Descent~\cite{robbins1951stochastic} or Adam~\cite{kingma2015adam} are typically employed. Therefore, our method simultaneously optimizes the camera hardware with a derivative-free optimizer and the trainable perception tasks with gradient-based optimizers. However, perception tasks that are not trainable, such as extracting Oriented FAST and Rotated BRIEF (ORB) features~\cite{rublee2011orb}, are not jointly optimized.

\smallskip \noindent \textbf{Camera Parameters}\hspace{2pt} Our optimizer can handle the optimization of all parameters captured in the camera simulation, e.g.~those outlined in Sec.~\ref{sec:camera_model}, which can be both continuous and discrete. For instance, parameters related to optics and image sensors can be optimized as continuous variables if there are no manufacturing constraints on new optics/sensors. Alternatively, they can be selected from existing lens/sensor catalogs, allowing manufacturing and availability to be considered.

\smallskip \noindent \textbf{Discrete Variable Optimization}\hspace{2pt} Our approach offers two schemes to optimize the discrete camera variables: \textit{fully discrete} and \textit{quantized continuous}. In the fully discrete scheme, we optimize the parameter $x$, representing the discrete parameter in its available values, by constraining the mutation stage of the genetic algorithm to ensure that only available values of this variable are used.
\begin{figure*}
      \centering
      \includegraphics[width=\linewidth]{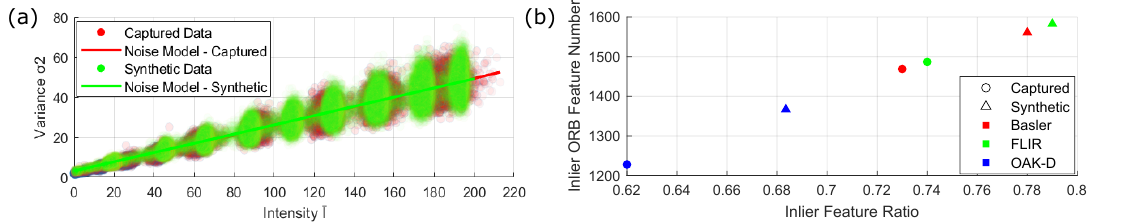}
      \caption{Comparison of captured and synthetic images in terms of (a) variance in pixel intensities and (b) perception task performance. In (a), despite differences in color intensities due to manufacturing variations of the test target, the variances of pixel values in synthetic images match those in captured images, validating the accuracy of our noise model. In (b), the ranking of camera performance in our simulation aligns with physical cameras, and the differences in their performance between captured and synthetic images are consistent.}
      \label{fig:sim_val}
\end{figure*}

In the quantized continuous scheme, we adopt the ``quantized continuous variables'' method introduced in \cite{cote2023differentiable}. Here, in each iteration, the discrete parameter $x$ can freely change as a continuous variable from its current best value obtained in the previous iteration. However, it is then replaced with the closest value from its available range:
\begin{equation}\label{eqn:quant_cont}
x^* = \argmin_k ||x - x_{k}||_2^2,
\end{equation}
where $x^*$ is the parameter retained from its available range, $x$ is the variable obtained from the optimization process, and $x_{k}$ represents the $k$-th parameter in its available values.

A limitation of the fully discrete scheme arises when a variable $x$ encompasses multiple interconnected parameters. For instance, if $x$ represents available image sensors, it includes parameters such as width, height, and pixel size. Selecting $x$ categorically does not leverage the relationships between these parameters. The quantized continuous scheme addresses this by optimizing all parameters within $x$ freely and then replaced by it with the closest categorical $x$. Therefore, the fully discrete scheme is suitable for independent parameters like the number of cameras, while the quantized continuous scheme is beneficial for parameters with interdependencies like image sensors.

\smallskip \noindent \textbf{Fitness Function}\hspace{2pt} The fitness function is constructed based on the performance of tasks, which is used to optimize the camera parameters using the derivative-free optimizer. We demonstrate the construction of the fitness function incorporating various perception tasks in our experiments.


\section{Experiments}
\label{sec:experiments}
We apply our proposed method to two design problems for demonstration: designing a stereo camera with two components for depth estimation on a vehicle and designing a monocular camera for multiple tasks on a Mixed Reality (MR) device. Additionally, we validate our camera simulation approach by comparing it with physical cameras. Additional details and results are provided in the supplement.

\smallskip \noindent \textbf{Assumptions}\hspace{2pt} (1) Objects' distances to the camera exceed the camera's hyperfocal distance so that the DOF is safely neglected. (2) The lens of our camera is free of geometric distortion and chromatic aberration.

\smallskip \noindent \textbf{Noise Synthesis}\hspace{2pt} We apply the affine noise model calibrated using a FLIR Flea3 Camera~\cite{FLIR2017} with a Sony IMX172 sensor to all synthetic images, except the simulations of off-the-shelf cameras used in Sec.~\ref{sec:simulator_val} and ~\ref{sec:mono_mr}, which are calibrated using their corresponding noise model.

\subsection{Simulator Validation}
\label{sec:simulator_val}
To validate our simulator, we compare synthetic images from our simulator with those captured by physical cameras in terms of image statistics and task performance.

\smallskip \noindent \textbf{Image Statistics}\hspace{2pt} We use a test target with linear greyscale colorbars in a controlled illumination environment, capturing images with the FLIR Flea3 Camera~\cite{FLIR2017}. The same test target is then recreated in our simulator, maintaining consistent illumination, camera position, and parameters. Fig.~\ref{fig:sim_val}(a) shows intensity variances versus pixel intensities for both captured and synthetic images, with good alignment validating the accuracy of our noise model. The slight differences in mean intensity for each greyscale bar are caused by printer variations during manufacturing.

\smallskip \noindent \textbf{Perception Task Performance}\hspace{2pt} We further validate the cameras' performance on perception tasks in our simulation, which serves as the evaluation method in Sec.~\ref{sec:mono_mr}. In this experiment, we focus on a feature extraction task using the widely applied ORB feature extractor~\cite{rublee2011orb}.

We use a test target from \cite{dansereau2019liff} with features of varying scales and depths, set against a texture-less background. Ten frames are captured with constant translational motion, ensuring consistent illumination, camera parameters, and relative positioning between real and virtual experiments. Fig.~\ref{fig:sim_val} (b) shows the number of inlier features (correctly matched points across consecutive frames) along with the ratio of inliers to total features (inliers plus outliers). 

The comparison includes three robotic/machine vision cameras: the RGB camera of the Luxonis OAK-D Pro Wide~\cite{OAKD2021}, the FLIR Flea3~\cite{FLIR2017}, and the Basler Dart DaA1280-54uc~\cite{Basler2024}. Specific noise models for these cameras, calibrated as described in Sec.~\ref{sec:camera_model}, are applied in our simulator. The feature extraction task serves as a reliable measure of performance consistency between physical and simulated environments.

Fig.~\ref{fig:sim_val} (b) shows that synthetic images from our simulator yield more inlier features and a higher inlier ratio due to quality reduction in the physical test target from the manufacturing process. However, the relative performance of the cameras, crucial for optimization, remains consistent between simulated and physical settings. This validates that our simulations accurately reflect real-world camera performance and effectively evaluate camera designs.

\subsection{Stereo Camera on Autonomous Vehicle}
\label{sec:depth_est}
We apply our method to design a stereo camera on an autonomous vehicle for the task of depth estimation.

\smallskip \noindent \textbf{Environment and Data}\hspace{2pt} This experiment is conducted in CARLA~\cite{dosovitskiy2017carla}, based on UE 4. The stereo camera is mounted to a car that moves automatically. The environment is configured with constant illumination, and motion blur enabled due to the moving platform. Images are captured during both training and testing. In training, each camera configuration captures one image per step, totalling 1000 images over 1000 steps. For testing, each camera design captures 1500 images to evaluate performance. Training and testing are performed on different urban outdoor maps.

\begin{figure}[t]
        \centering
        \includegraphics[width=\linewidth]{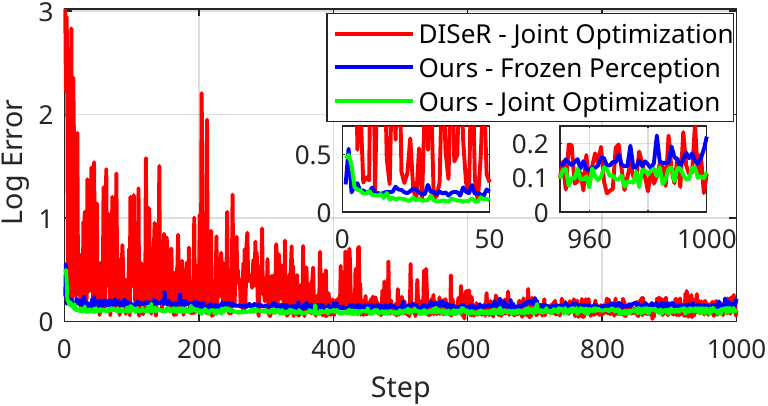}
        \caption{Training curves comparing the design of cameras using our method with and without joint optimization, and curve of DISeR~\cite{klinghoffer2023diser}, are plotted. Zoomed-in windows from 0 to 50 and 950 to 1000 timesteps are provided for visualization. The curves demonstrate that our method with joint optimization achieves superior task performance with fewer steps and smoother behaviour.}
\label{fig:curves}
\end{figure}

\smallskip \noindent \textbf{Design Space}\hspace{2pt} We optimize the baseline ($b \in [0.01 \text{m}, 3 \text{m}]$) and horizontal FOV ($fov \in [50^{\circ}, 120^{\circ}$]) as continuous variables. We use two cameras with fixed sensor sizes of 1.536 mm$\times$0.768 mm, a pixel size of 1.55 $\mu$m, and a mounting height of 2 m, positioned forward-facing with no slant.

\smallskip \noindent \textbf{Stereo Matching Network}\hspace{2pt} We use the PSMNet~\cite{chang2018pyramid} to predict the disparity map, which is pretrained on the KITTI Stereo 2015 dataset~\cite{Menze2015ISA, Menze2018JPRS} and subsequently fine-tuned with the optimization of camera parameters using its loss function ($l_{disparity}$, Smooth L1 Loss), and the maximum disparity for this model is capped as 192 following \cite{chang2018pyramid}. We use an Adam optimizer~\cite{kingma2015adam} with a batch size of 4 and a learning rate of $1\cdot10^{-3}$ to train the network.

\smallskip \noindent \textbf{Fitness Function}\hspace{2pt} We use the inverse of the log error between the predicted and ground-truth disparity maps (normalized by baselines), which emphasizes the depth prediction accuracy in both large and small distances.

\smallskip \noindent \textbf{Derivative-Free Optimization}\hspace{2pt} The genetic algorithm uses 5 solutions per generation, retaining the top 2 for the next iteration. The top 3 generate offspring via uniform crossover, followed by mutation with a random factor (0.8 to 1.2) and an addition value (-5 to 5 for FOV, -0.2 to 0.2 for baseline). Hyperparameters are empirically chosen, and optimization parameters are randomly initialized. See the supplement for comparisons of hyperparameters and initialization.

\smallskip \noindent \textbf{Results}\hspace{2pt} Tab.~\ref{tab:results_stereo} details the performance of stereo cameras designed by our method, compared with two off-the-shelf models, the Intel RealSense D450~\cite{Intel2024} and ZED 2i~\cite{StereoLabs}, as well as a camera designed with the RL method DISeR~\cite{klinghoffer2023diser}. We also compare results from jointly optimizing camera design and perception tasks against optimizing camera design alone while fixing perception model parameters (pretrained with 2500 images from CARLA using 50 camera configurations). \textcolor{green}{$\bigcdot$} indicates optimized parameters and $\bigcdot$ indicates fixed parameters. Performance is evaluated using Average Log Error and Root Mean Square Error (RMSE) between estimated and ground-truth depths in meters.

Our camera design and DISeR perform best in terms of log error, which evaluates depth estimation accuracy across both short and long distances, and RMSE, which favors larger baselines for more accurate long-distance estimation as long-distance has a greater impact on this value. However, larger baselines struggle with short-distance accuracy due to limited overlapping FOVs and the perception model's maximum disparity setting. While increasing FOVs can improve overlap, it reduces accuracy by lowering disparity values. As a result, both our method and DISeR converge on small FOVs and moderate baselines to balance accuracy across all distances. See the supplementary material for qualitative results and analysis.

Fig.~\ref{fig:curves} compares the training curves for our joint optimization method, our method with fixed perception parameters, and the RL method. Our method converges faster than DISeR by directly optimizing camera parameters, whereas DISeR optimizes a policy network with many parameters to predict them. The RL method shows abrupt changes during training, as all network parameters are updated in each iteration, leading to significant fluctuations in camera parameters and performance. In contrast, the genetic algorithm makes smaller parameter adjustments, resulting in smoother performance changes. Our method converges with 45 steps in 2 min and completes 1000 steps in 38 minutes with an NVIDIA RTX4070 GPU, while DISeR takes 700 steps (67 min) to converge and 97 minutes to complete 1000 steps.

\begin{table}
\centering
\caption{
Depth estimation performance of stereo cameras designed using our method, the RL method (DISeR~\cite{klinghoffer2023diser}), and two off-the-shelf cameras. The proposed method incorporating joint optimization of camera parameters and perception tasks shows similar performance as DISeR at a fraction of the training time.
}
\resizebox{\columnwidth}{!}{%
\begin{tabular}{ccccc}
\hline
\multirow{3}{*}{Camera} & \multicolumn{2}{c}{Camera Parameters} & \multicolumn{2}{c}{Performance}\\
\cline{2-5}
& Baseline & Horizontal FOV & Log Error & RMSE\\
& $b$ (m) & $fov$ (\textdegree) & $\downarrow$ & $\downarrow$\\
\hline
RealSense~\cite{Intel2024} & 0.095 $\bigcdot$ & 87 $\bigcdot$ & 0.39\textcolor{green}{$\bigcdot$} & 178.94\textcolor{green}{$\bigcdot$}\\
ZED~\cite{StereoLabs} & 0.12 $\bigcdot$ & 72 $\bigcdot$ & 0.29\textcolor{green}{$\bigcdot$} & 134.74\textcolor{green}{$\bigcdot$}\\
DISeR~\cite{klinghoffer2023diser} & 1.84\textcolor{green}{$\bigcdot$} & 50\textcolor{green}{$\bigcdot$} & 0.16 \textcolor{green}{$\bigcdot$} & \textbf{75.05}\textcolor{green}{$\bigcdot$}\\
Ours - Frozen & 1.41\textcolor{green}{$\bigcdot$} & 50 \textcolor{green}{$\bigcdot$} & 0.19 $\bigcdot$ & 111 $\bigcdot$\\
Ours - Joint & 1.6\textcolor{green}{$\bigcdot$} & 50 \textcolor{green}{$\bigcdot$} & \textbf{0.14} \textcolor{green}{$\bigcdot$} & 79.81 \textcolor{green}{$\bigcdot$}\\
\hline
\end{tabular}
}
\vspace{-0.05in}
\label{tab:results_stereo}
\end{table}

\begin{table*}
\centering
\caption{
We compared the parameters and performance of cameras designed with our method using both fully discrete and quantized continuous schemes alongside three robotic/machine vision cameras. Optimized parameters via our method are labeled with \textcolor{green}{$\bigcdot$}, covering all camera parameters, the object detection network in the joint optimization scheme, and mounting angles of off-the-shelf cameras. Our approach consistently designs cameras with higher performance in both scenarios. The quantized continuous scheme notably outperforms the fully discrete scheme, benefiting from its consideration of parameter interdependencies. Joint optimization of the object detector also enhances detection precision, while the performance of non-trainable tasks remains similar.
}
\resizebox{2\columnwidth}{!}{%
\begin{tabular}{cccccccccc}
\hline
\multirow{3}{*}{Scenario} & \multirow{3}{*}{Camera} &\multicolumn{4}{c}{Camera Parameters}& \multicolumn{4}{c}{Performance} \\
\cline{3-10}
&  & Pitch Angle & Focal Length & Sensor Size & Pixel Size & Obstacle Detect. & Object Detect. & Inlier Number & Inlier Ratio\\
& & $\theta$ ($^{\circ}$) & $f$ (mm) & $w\times h$  (mm) & $p$ ($\mu$m) & Accuracy $\uparrow$ & AP $\uparrow$ & $\uparrow$ & $\uparrow$\\
\hline
\multirow{7}{*}{Day} & OAK-D~\cite{OAKD2021} & -20.04 \textcolor{green}{$\bigcdot$} & 2.75 $\bigcdot$ & 6.29$\times$4.71 $\bigcdot$ & 1.55 $\bigcdot$ & 1 & 0.37 \textcolor{green}{$\bigcdot$} & 115 & 0.07\\
& FLIR~\cite{FLIR2017} & -23.28 \textcolor{green}{$\bigcdot$} & 3.6 $\bigcdot$ & 6.2$\times$4.65 $\bigcdot$ & 1.55 $\bigcdot$ & 1 & 0.37 \textcolor{green}{$\bigcdot$} & 77 & 0.05\\
& Basler~\cite{Basler2024} & -25.90 \textcolor{green}{$\bigcdot$} & 3.6 $\bigcdot$ & 4.8$\times$3.6 $\bigcdot$ & 3.75 $\bigcdot$ & 1 & 0.23 \textcolor{green}{$\bigcdot$} & 131 & 0.08\\
\cline{2-10}
& \multirow{2}{*}{Ours - Fully Discrete} & -27.87 \textcolor{green}{$\bigcdot$} & 4.01 \textcolor{green}{$\bigcdot$} & 8.45$\times$6.76 \textcolor{green}{$\bigcdot$} & 6.6 \textcolor{green}{$\bigcdot$} & 1 & 0.43 $\bigcdot$ & 191 & \textbf{0.13}\\
& & -24.69 \textcolor{green}{$\bigcdot$} & 3.77 \textcolor{green}{$\bigcdot$} & 8.45$\times$6.76 \textcolor{green}{$\bigcdot$} & 6.6 \textcolor{green}{$\bigcdot$} & 1 & 0.51 \textcolor{green}{$\bigcdot$} & 189 & 0.12\\
\cline{2-10}
& \multirow{2}{*}{Ours - Quantized Continuous} & -21.86 \textcolor{green}{$\bigcdot$} & 2.99 \textcolor{green}{$\bigcdot$} & 8.45$\times$6.76 \textcolor{green}{$\bigcdot$} & 6.6 \textcolor{green}{$\bigcdot$} & 1 & 0.46 $\bigcdot$ & \textbf{230} & \textbf{0.13}\\
& & -26.34 \textcolor{green}{$\bigcdot$} & 2.88 \textcolor{green}{$\bigcdot$} & 7.31$\times$5.58 \textcolor{green}{$\bigcdot$} & 4.5 \textcolor{green}{$\bigcdot$} & 1 & \textbf{0.64} \textcolor{green}{$\bigcdot$} & 224 & \textbf{0.13}\\
\hline
\multirow{7}{*}{Night} & OAK-D~\cite{OAKD2021} & -19.72 \textcolor{green}{$\bigcdot$} & 2.75 $\bigcdot$ & 6.29$\times$4.71 $\bigcdot$ & 1.55 $\bigcdot$ & 1 & 0.34 \textcolor{green}{$\bigcdot$} & 117 & 0.06\\
& FLIR~\cite{FLIR2017} & -22.71 \textcolor{green}{$\bigcdot$} & 3.6 $\bigcdot$ & 6.2$\times$4.65 $\bigcdot$ & 1.55 $\bigcdot$ & 1 & 0.36 \textcolor{green}{$\bigcdot$} & 69 & 0.03\\
& Basler~\cite{Basler2024} & -25.83 \textcolor{green}{$\bigcdot$} & 3.04 $\bigcdot$ & 4.8$\times$3.6 $\bigcdot$ & 3.75 $\bigcdot$ & 1 & 0.16 \textcolor{green}{$\bigcdot$} & 122 & 0.07\\
\cline{2-10}
& \multirow{2}{*}{Ours - Fully Discrete} & -23.65 \textcolor{green}{$\bigcdot$} & 2.61 \textcolor{green}{$\bigcdot$} & 7.2$\times$5.4 \textcolor{green}{$\bigcdot$} & 4.5 \textcolor{green}{$\bigcdot$} & 1 & 0.37 $\bigcdot$ & \textbf{179} & \textbf{0.11}\\
&&  -23.66 \textcolor{green}{$\bigcdot$} & 3.10 \textcolor{green}{$\bigcdot$} & 7.2$\times$5.4 \textcolor{green}{$\bigcdot$} & 4.5 \textcolor{green}{$\bigcdot$} & 1 & 0.42 \textcolor{green}{$\bigcdot$} & 170 & \textbf{0.11}\\
\cline{2-10}
& \multirow{2}{*}{Ours - Quantized Continuous} & -29.86 \textcolor{green}{$\bigcdot$} & 3.33 \textcolor{green}{$\bigcdot$} & 8.45$\times$6.76 \textcolor{green}{$\bigcdot$} & 6.6 \textcolor{green}{$\bigcdot$} & 1 & 0.37 $\bigcdot$ & 165 & \textbf{0.11}\\
& & -22.58 \textcolor{green}{$\bigcdot$} & 3.49 \textcolor{green}{$\bigcdot$} & 14.48$\times$9.94 \textcolor{green}{$\bigcdot$} & 9 \textcolor{green}{$\bigcdot$} & 1 & \textbf{0.57} \textcolor{green}{$\bigcdot$} & 172 & \textbf{0.11}\\
\hline
\end{tabular}
}
\vspace{-0.05in}
\label{tab:results}
\end{table*}

\subsection{Monocular Camera on Mixed Reality Headset}
\label{sec:mono_mr}
In our second experiment, we apply our method to design a monocular RGB camera for an MR headset. Object detection, obstacle avoidance, and feature extraction for 3D reconstruction are selected as examples of tasks as they are essential for most MR devices. However, other tasks can be added depending on the target application.

\smallskip \noindent \textbf{Environment and Data}\hspace{2pt} We establish an indoor environment in UE 5 with 10 object classes. Floorplans and object locations are randomly generated to introduce variability and the camera is mounted on an auto-agent acting as a user. During training, each camera design captures 500 images, indicating 10000 images in total as we optimize for 20 generations, 10 solutions per generation. Testing is conducted with 1000 images using different scene configurations.

\smallskip \noindent \textbf{Design Space}\hspace{2pt} We focus on designing geometric parameters that determine the camera's FOV and photometric parameters affecting resolution, crucial for user mobility and objects' effective resolution. For FOV, we optimize the mounting angle in pitch direction ($\theta \in [-30^{\circ}, 30^{\circ}]$, focal length ($f \in [1 \text{mm}, 20 \text{mm}]$), and image sensor dimensions (width $w$ and height $h$). The number of pixels is decided by the sensor's pixel size ($p$) and its dimensions. The camera's height varies randomly between 1 m to 2 m per training step to ensure robustness across different user heights.

We restrict our design to readily available sensors by optimizing the image sensor ($i$) as a categorical variable, selecting from a catalog of 43 commercial CMOS image sensors from five manufacturers ($i_{c1}$, $i_{c2}$,..., $i_{c43}$). Each sensor ($i$) comprises a set of sensor-related parameters ($w$, $h$, and $p$). We compare two optimization techniques for discrete variables, treating $i$ as fully discrete and using the quantized continuous approach. Our catalog predominantly features 28 Sony sensors, reflecting their widespread use in machine vision cameras, and our noise model is more likely to generalize well across sensors from the same manufacturer. Hence, we also used a Sony sensor to calibrate.

\smallskip \noindent \textbf{Obstacle Avoidance}\hspace{2pt} To assess the camera’s ability to detect obstacles affecting user mobility, we place thresholds at room entrances in our virtual environment. These low-height obstacles highlight the importance of an appropriate FOV and mounting angle. We focus on low-height obstacles because taller ones are constrained by object and feature detection tasks. Visibility to the camera is determined by whether the obstacle appears in rendered images. Additionally, the auto-agent is programmed to react when stepping on a threshold, regardless of its visual rendering.

\smallskip \noindent \textbf{Feature Extraction}\hspace{2pt} We extract ORB features~\cite{rublee2011orb} and employ the Brute-Force Matcher in OpenCV~\cite{itseez2015opencv} to match features across consecutive frames, we then apply RANSAC~\cite{fischler1981random} to find inlier features while estimating transformation matrix between frames.
\begin{figure*}
      \centering
      \includegraphics[width=\linewidth]{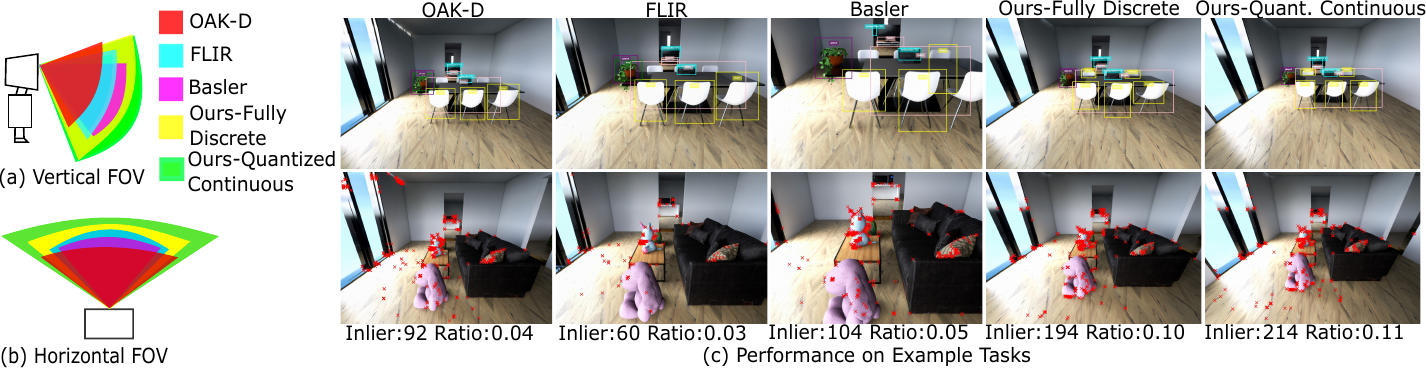}
      \caption{The vertical (a) and horizontal (b) FOVs of cameras optimized by our method with joint optimization, alongside those designed by humans under the daytime scenario, and example evaluation (c). Our method designs a camera with the largest FOV, enabling the capture of all obstacles and objects, extracting more features while maintaining sufficient resolution for effective object and feature detection.}
      \label{fig:Results_Demo}
\end{figure*}

\smallskip \noindent \textbf{Object Detector}\hspace{2pt} We utilize a Faster R-CNN~\cite{ren2015faster} object detector with a ResNet-50~\cite{he2016deep} backbone, pretrained on 2000 images generated by our simulator and then fine-tuned alongside the camera parameters. Training employs an Adam optimizer~\cite{kingma2015adam} with a batch size of 8 and a learning rate of $1\cdot10^{-4}$, which decays by 0.5 every 5 steps during both pre-training and fine-tuning.

\smallskip \noindent \textbf{Fitness Function}\hspace{2pt} The fitness function combines three perception tasks. For obstacle avoidance, it calculates the ratio of the number of obstacles seen by the camera ($o_{seen}$) to the total number of obstacles ($o_{total}$) in the user's path. The object detector is trained using its loss function ($l_{OD}$), and we take the Average Precision (AP) with a 0.5 Intersection-over-Union (IoU) threshold as a term in the fitness function. Feature extraction contributes through the number of inlier features ($n_{inlier}$) and the ratio of inlier features to total features ($n_{total}$), emphasizing both the quantity and accuracy of detected features. Thus, the total fitness function is 
\begin{equation}\label{eqn:rgb_fitness}
\begin{aligned}
F = \lambda_{feature} (\lambda_{inlier} n_{inlier} + \lambda_{ratio}\frac{n_{inlier}}{n_{total}})\\ + \lambda_{OD}\text{AP}@0.5IoU + \lambda_{obstacle} \frac{o_{seen}}{o_{total}},
\end{aligned}
\end{equation}
where we balance the weights of all terms by setting $\lambda_{inlier}$ to 0.0025 (inlier in an image is typically 100-250), $\lambda_{ratio}$ to 0.5, $\lambda_{obstacle}$, $\lambda_{OD}$, and $\lambda_{feature}$ to 1 as we consider these tasks equally significant.

\smallskip \noindent \textbf{Derivative-Free Optimization}\hspace{2pt} We optimize the camera parameters using the genetic algorithm over 20 generations with 10 solutions per generation. A uniform crossover is applied using the top 5 solutions to produce offspring, while the top 3 solutions are reused. The mutation process remains consistent with the previous experiment, except the addition value is randomly selected between -3 and 3. These hyperparameters are chosen empirically. Each solution involves collecting 500 images to evaluate both the feature extraction and obstacle (threshold) avoidance tasks.

\smallskip \noindent \textbf{Results}\hspace{2pt} We present the optimized set of camera parameters obtained through our approach and evaluate the camera's performance in Tab.~\ref{tab:results}, considering obstacle detection accuracy, AP score, average number of inlier ORB features across consecutive frames, and ratio of average inlier features to total features extracted. Additionally, we report optimized parameters under two application scenarios: daytime operation in a well-illuminated simulation environment (20 lux) with a lower baseline camera gain (5 dB), and nighttime operation in low-light conditions (2 lux) with a higher baseline camera gain (15 dB). We observed that different application scenarios led to distinct camera designs. Our approach designed a camera with a larger pixel size for nighttime applications, which is expected as a sensor with larger pixels delivers higher SNR since larger pixels gather more light. Hence, sensors with larger pixels require lower gain to capture images with the comparable measured intensity compared to sensors with smaller pixels.

We compare our optimized camera with three human-designed robotic/machine vision cameras, OAK-D, FLIR, and Basler, used in previous experiments in our simulation. Specific noise models are applied to these cameras, with optimized mounting angles while others remain fixed due to configurability. Additionally, we report performance without joint optimization, where parameters of the object detection network are frozen. Fig.~\ref{fig:Results_Demo} illustrates FOVs and example evaluations of our proposed cameras with joint optimization alongside robotic/machine vision cameras. Please refer to the supplementary material for more visualisations.

The results show performance improvements across all tasks with our method, while lower performance is observed under night scenarios due to reduced SNR. Fully discrete optimization schemes show lower performance, indicating the importance of considering parameter interdependencies. Similarly, freezing parameters of the perception model results in reduced performance. The performance of obstacle avoidance is always perfect because we optimized the pitch angle of all the cameras, including the off-the-shelf ones.

We note that the RL method was developed for and demonstrated on single-frame tasks~\cite{klinghoffer2023diser}. Both feature matching and obstacle avoidance are episodic, multi-frame tasks requiring consecutive images for precise performance evaluation, and this data collection process is time-consuming. The RL method demands significantly more training steps and image data than ours, applying it to episodic tasks requires impractically long optimization times that prevent us from making a direct comparison. Hence, it is not compared in this experiment.

\section{CONCLUSION}
We presented a novel end-to-end approach that combines derivative-free and gradient-based optimizers to co-design cameras with perception tasks efficiently. Utilizing UE and an affine noise model, we constructed a camera simulator and validated its accuracy against physical cameras. Our method handles continuous, discrete, categorical camera parameters, and advances a quantized continuous approach for discrete variables to consider their interdependencies. We believe this work can be generalized easily and is an important step toward principled and automated camera design for autonomous systems that account for the interdependency between cameras and the algorithms that interpret them. For future work, we aim to develop a task-driven control algorithm that dynamically adjusts camera parameters, such as exposure settings, in an online manner.

\smallskip \noindent \textbf{Acknowledgements}\hspace{2pt} We would like to thank both ARIA Research Pty Ltd and the Australian government for their funding support via a CRC Projects Round 11 grant.

{\small
\bibliographystyle{ieee_fullname}
\bibliography{egbib}
}

\clearpage \appendix \input{Supplementray}

\end{document}

%% file: Supplementray.tex
\begin{figure}[h]
      \centering
      \includegraphics[width=\linewidth]{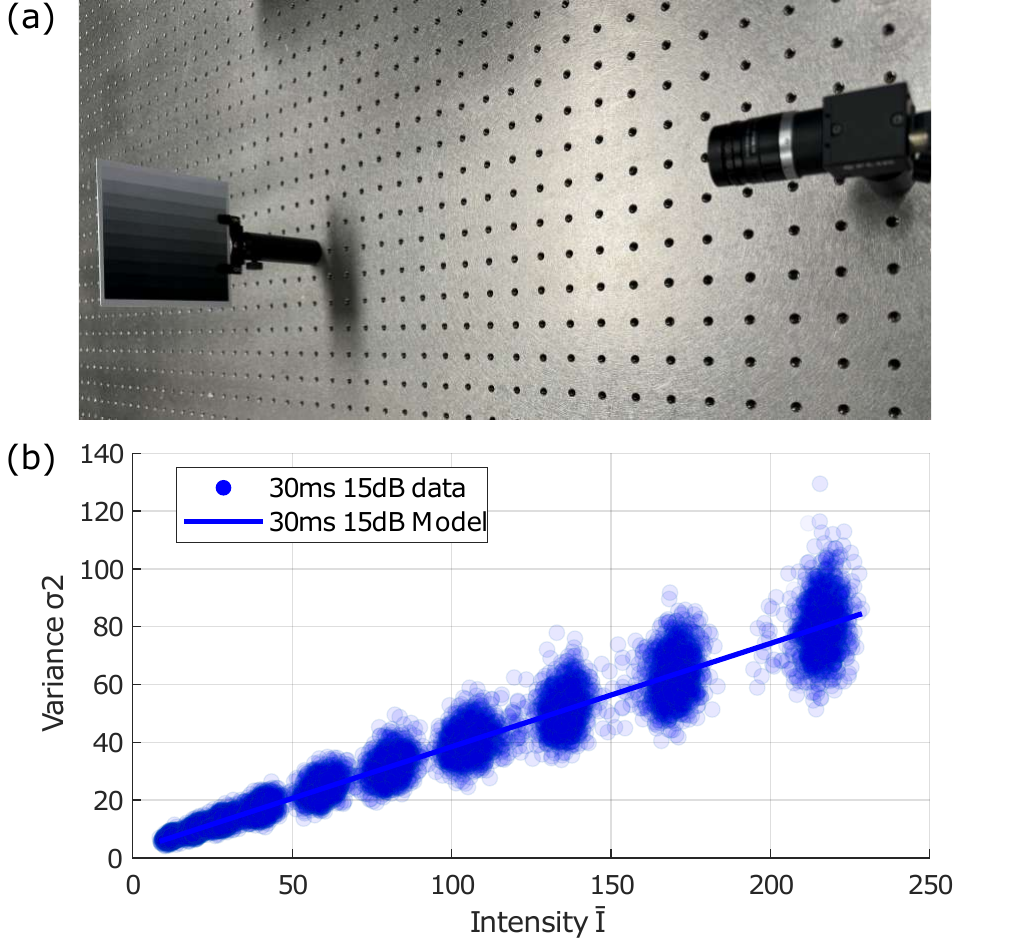}
      \caption{We calibrate the affine noise model~\cite{foi2009clipped} with a FLIR Flea3 Camera~\cite{FLIR2017} using a greyscale test target with colorbars. The camera is configured to have a 30 ms exposure time and 15 dB gain. The calibration setup is illustrated in (a) and the plot of the pixel variances against mean intensities is displayed in (b).}
      \label{fig:noise_cal}
\end{figure}
\section{Code}
Our code is available on our project page at \href{https://roboticimaging.org/Projects/TaCOS/}{\textcolor{magenta}{https://roboticimaging.org/Projects/TaCOS/}}. We include the implementation of our camera design method for both the stereo camera and monocular camera design experiments, the source code and guidance for creating the indoor virtual environment used in the monocular camera design experiment, and the catalog of commonly available image sensors that we collected.

\section{Additional Details on Noise Synthesis}
We provide additional details on our image noise calibration and generalization method.

\begin{figure}
      \centering
      \includegraphics[width=\linewidth]{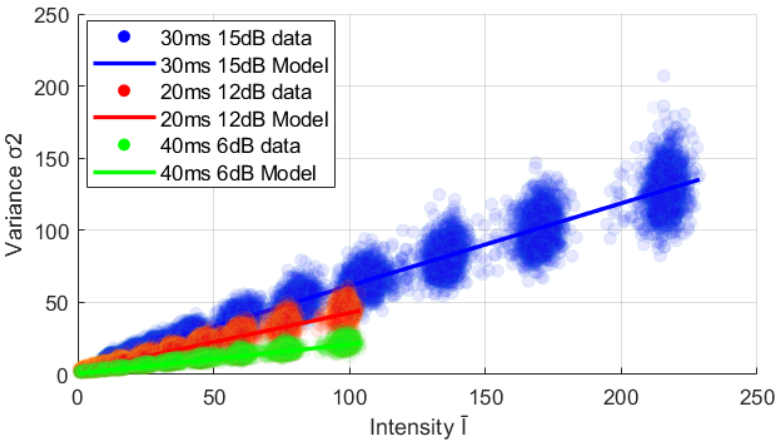}
      \caption{The noise model calibrated with 30 ms exposure time and 15 dB gain (blue) is generalized to the 20 ms exposure time and 12 dB gain (red), as well as 40 ms exposure time and 6 dB gain (green). The data on the plot are captured by the physical camera, the noise model for 30 ms exposure time and 15 dB gain are obtained from the data as in Fig.~\ref{fig:noise_cal} (b), while noise models for the latter two exposure settings are obtained via our generalization method. The plot shows that generalized noise models match the captured data accurately. The intensities for the latter two exposure settings are smaller, which is caused by lower exposure settings.}
      \label{fig:noise_genl}
\end{figure}

\subsection{Noise Model Calibration}
\label{sec:noise_cal}
We adopt the affine noise model~\cite{foi2009clipped} in this work. The noise model describes a linear relationship between the variance ($\sigma^2$) in image pixel intensities for different mean intensity values ($\Bar{I}$) in terms of constant thermal noise ($\sigma_t$) and intensity-varying photon noise ($\sigma_p^2 \Bar{I}$):
\begin{equation}\label{eqn:noise model}
\sigma^2 = \sigma_p^2 \Bar{I} + \sigma_{t}^2.
\end{equation}

Calibrating the noise model follows established methods~\cite{ratner2007illumination, liu2006noise, wang2021multiplexed}. In this work, we use a greyscale test target with colorbars containing uniformly distributed grey levels from fully white to fully black. With captured images of the test target, we determine mean intensities and variances for each pixel, using these values to fit the affine noise model defined in Eq.~\ref{eqn:noise model}.

We calibrate the noise model with a FLIR Flea3 Camera~\cite{FLIR2017} with a Sony IMX172 image sensor. The exposure time is set as 30 ms and the gain is set as 15 dB. Note that the dark current noise is safely neglected in this work as the exposure time used (30 ms) is relatively short. The setup of this calibration is demonstrated in Fig.~\ref{fig:noise_cal} and we display the plot of the obtained noise model in Fig.~\ref{fig:noise_cal} (b). 

\begin{figure*}
      \centering
      \includegraphics[width=\linewidth]{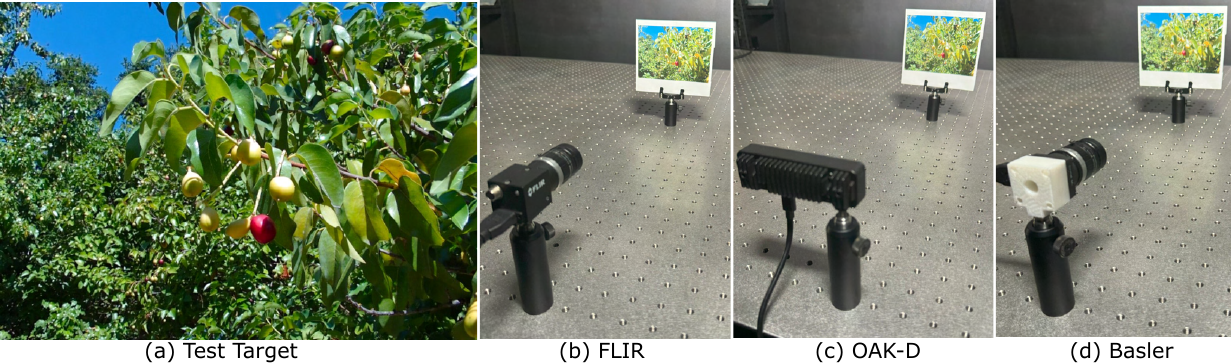}
      \caption{We adopt a test target (a) from \cite{dansereau2019liff} to validate our simulator in terms of the performance of feature extraction. We compare the relative performance of three off-the-shelf cameras, the experiment setups in the real world are shown in (b), (c), and (d), which are then duplicated in our virtual environment.}
      \label{fig:feature_val}
\end{figure*}

\subsection{Noise Model Generalization}
The noise model in Eq.~\ref{eqn:noise model} is calibrated using a specific image sensor and exposure setting. We expand the equation so that it can be generalized to different exposure settings and image sensors.

Consider the intensity ($I$) in Eq.~\ref{eqn:noise model} as the measured pixel intensity by the camera: $I = EG\phi$, where $E$ and $G$ are the exposure time and gain respectively, and $\phi$ is the scene radiance. Intensity changes due to exposure time setting are reflected in the measured intensity value, therefore, we generalize the noise model to other exposure and gain settings by multiplying the ratio of the new gain ($G$) and the calibrated gain ($G_0$) used in the noise calibration stage. Then we can transform Eq.~\ref{eqn:noise model} to obtain the noise model with new exposure settings. For the photon noise term, replacing $\Bar{I}$ in Eq.~\ref{eqn:noise model} with the new observed intensity and multiplying it with the gain ratio to scale the value of $\sigma_{p}^2$. For the thermal noise term, scaling $\sigma_{t}^2$ with the second order square of the ratio between the new gain and the calibrated gain, the noise model becomes Eq.~2 of the main paper.

In Fig.~\ref{fig:noise_genl}, we show the generalization of the calibrated noise model using 30 ms exposure time and 15 dB gain to two other exposure settings with the same camera, which are 20 ms exposure time with 12 dB gain and 40 ms exposure time and 6 dB gain. The data on the plot are captured with the camera, while the noise models are obtained with our derived generalization equation, which is Eq. 2 of the main paper. This experiment validates both the noise calibration and the generalisation of the noise model.

\section{Additional Details on Simulator Validation}
We provide implementation details for the simulator validation experiments. The results of the experiments are illustrated in Fig.~3 of the main paper.
\begin{figure}[t]
        \centering
        \includegraphics[width=\linewidth]{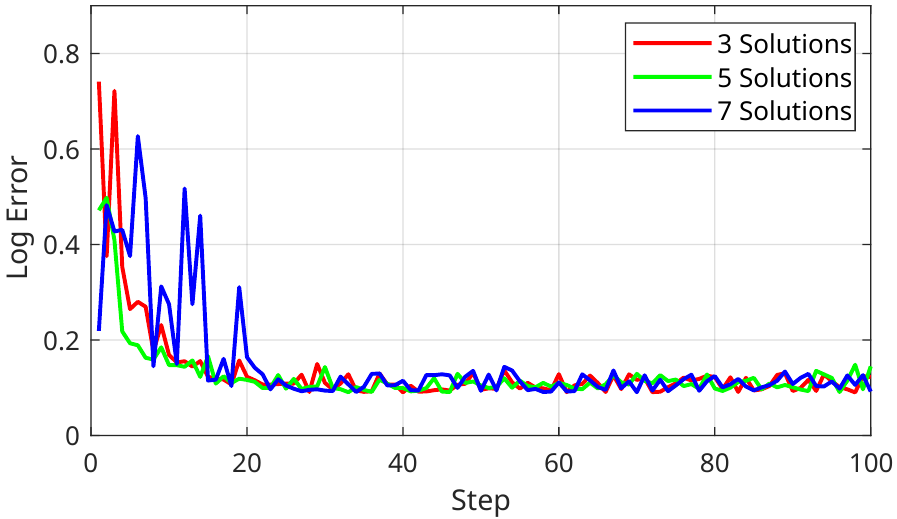}
        \caption{Training curves of 3 different solution numbers per generation, which are 3 solutions per generation, 5 solutions per generation (used in the main result), and 7 solutions per generation. The training curves indicate that all 3 settings can achieve a similar final performance, where the setting of 5 solutions per generation gives the smoothest and fastest convergence as it provides a search space that is more diverse compared to 3 solutions but also less to explore compared to 7 solutions. The experiment runs for 1000 steps but we only display the first 100 steps for visualization.}
\label{fig:ga_solnum}
\end{figure}
\begin{figure}[t]
        \centering
        \includegraphics[width=\linewidth]{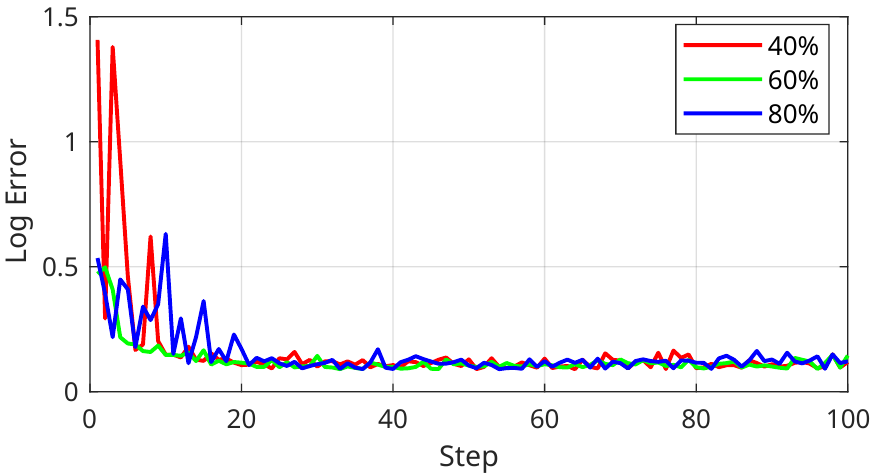}
        \caption{Training curves of 3 percentages of the population for offspring generation, which are 40\% of the population, 60\% of the population (used in the main result), and 80\% of the population. The curves show that using 60\% of solutions provides the fastest and smoothest convergence by balancing between exploration and exploitation, which is significantly beneficial for faster convergence. The experiment runs for 1000 steps but we only display the first 100 steps for visualization.}
\label{fig:parent_num}
\end{figure}
\begin{table}
\centering
\caption{
Comparison of the optimized camera parameters for the depth estimation task using the genetic algorithm with 3 initializations, including initialising all the parameters at their smallest values in the design space, initializing all the parameters at their largest values in the design space, and initializing all the parameters at random values within in the design space, which is the method taken in the main paper. The results show that the optimization is insensitive to initialization.
}
\resizebox{\columnwidth}{!}{%
\begin{tabular}{ccccc}
\hline
\multirow{3}{*}{Initialization} & \multicolumn{2}{c}{Camera Parameters} & \multicolumn{2}{c}{Performance}\\
\cline{2-5}
& Baseline & Horizontal FOV & Log Error & RMSE\\
& $b$ (m) & $fov$ (\textdegree) & $\downarrow$ & $\downarrow$\\
\hline
Smallest & 1.67\textcolor{green}{$\bigcdot$} & 50 \textcolor{green}{$\bigcdot$} & 0.14\textcolor{green}{$\bigcdot$} & 80.7\textcolor{green}{$\bigcdot$}\\
Largest & 1.64 \textcolor{green}{$\bigcdot$} & 50 \textcolor{green}{$\bigcdot$} & 0.14 \textcolor{green}{$\bigcdot$} & 78.05\textcolor{green}{$\bigcdot$} \\
Random & 1.6\textcolor{green}{$\bigcdot$} & 50 \textcolor{green}{$\bigcdot$} & 0.14 \textcolor{green}{$\bigcdot$} & 79.81 \textcolor{green}{$\bigcdot$}\\
\hline
\end{tabular}
}
\vspace{-0.05in}
\label{tab:initial}
\end{table}
\begin{figure}[t]
        \centering
        \includegraphics[width=\linewidth]{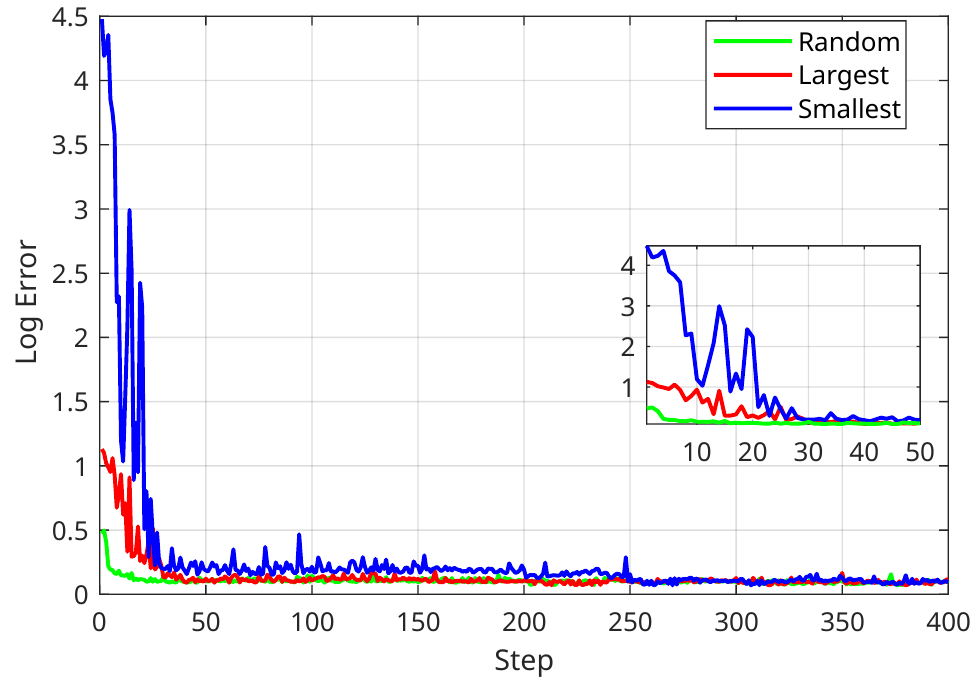}
        \caption{Training curves of 3 different initialization of camera parameters for the depth estimation task used by the genetic algorithm, which are starting from their smallest values, largest values, and random values within their design space. The curves show that the initialization does not affect the final performance but starting from random values results in faster convergence. The experiment runs for 1000 steps but we only display the first 400 steps for visualization.}
\label{fig:ga_init}
\end{figure}

\subsection{Image Statistics}
\label{sec:image_stat}
This experiment reuses the same test target and the FLIR camera as the noise level calibration experiment shown in Fig.~\ref{fig:noise_cal} (a). In this experiment, we strictly control the distance between the test target and the camera to 50 cm, and the illumination level at the test target as 2000 lux (measured with a light meter) using an LED panel light. The light is placed above the camera with an 80 cm distance and an angle of approximately 25\textdegree, facing downward to the test target. 

The same setup is then duplicated in Unreal Engine (UE), which is the simulator used in our experiments. In UE, the scene capture camera is set to have the same focal length, sensor size, pixel number, exposure time, and aperture size as the physical camera. However, we manually tune the ISO setting in the simulator to achieve the same brightness level as the physical camera. The rendered images are then applied with the noise model calibrated with the physical camera.

\subsection{Perception Task Performance}
\label{sec:validation_task}
To validate our simulator in the performance of extracting Oriented FAST and Rotated BRIEF (ORB)~\cite{rublee2011orb} features, we adopt a test target used in \cite{dansereau2019liff}, displayed in Fig.~\ref{fig:feature_val} (a), that is suitable for feature extraction. We use the same illumination setup in this experiment as described in Sec.~\ref{sec:image_stat}. This experiment was conducted with the RGB camera of the Luxonis OAK-D Pro Wide camera, the FLIR Flea3 Camera, and the Basler Dart DaA1280-54uc camera for comparison, which are shown in Fig.~\ref{fig:feature_val} (b), (c), and (d). In addition, the background of this experiment is textureless to avoid additional features, and the test target is moved and captured at 10 locations along the same horizontal line to simulate a translational motion for feature matching and determining the number of inlier features, motion blur is not considered for this experiment due to a short exposure time.

In our UE simulator, we also duplicate the setup in the physical experiment. Similarly, we configure the scene capture camera to have the same focal lengths, sensor sizes, pixel numbers, exposure times, and aperture sizes as the three physical cameras, and we manually calibrate the ISO settings to match their gain values. We calibrate the noise models for these three cameras individually for this experiment and apply their noise models to the renders. The noise model calibration method follows Sec.~\ref{sec:noise_cal}.

\section{Genetic Algorithm Implementation}
\subsection{Hyperparameter Selection}
\label{sec:hyperparam}
\smallskip \noindent \textbf{Population Size} The number of solutions per generation is selected empirically based on the number of parameters to optimize. For example, we only optimize 2 camera parameters in the stereo camera design example so that we choose a relatively small solution number (5 solutions per generation), and for a more complex problem like the monocular camera design example, we use a larger solution number (10 solutions per generation). This scheme is selected since more complex design problems generally require more diverse solutions to search through a larger search space. However, a larger number of solutions takes longer to converge since a larger search space is explored. Conversely, a smaller solution number gives a less diverse search space and encounters the issue of local optima. Hence, we empirically select the number of solutions in this work as illustrated in Fig.~\ref{fig:ga_solnum}, in which we compare the training curves of the depth estimation experiment with 3 different numbers of solutions per generation.

\begin{figure*}
      \centering
      \includegraphics[width=\linewidth]{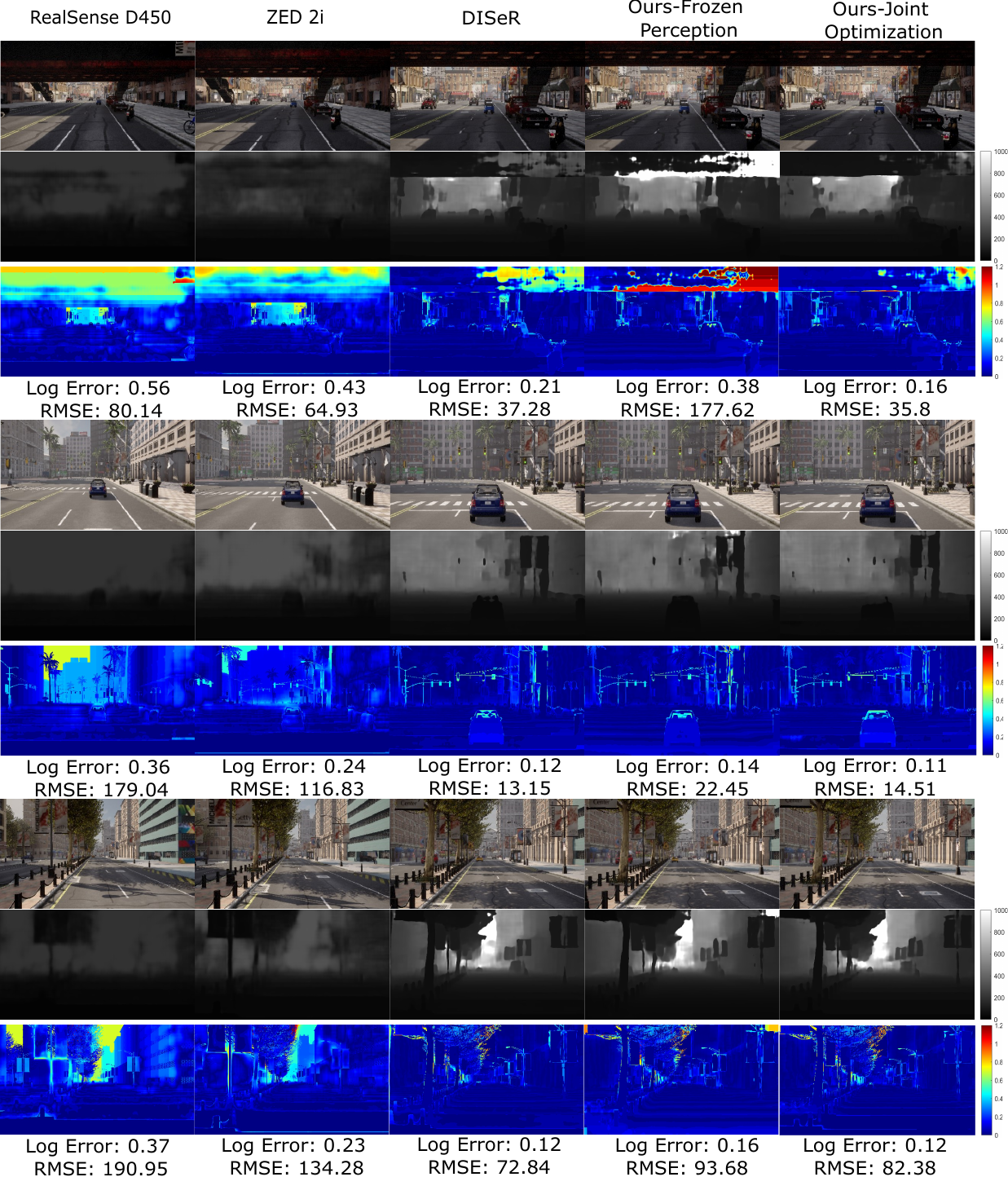}
      \caption{Comparison of captured left images, estimated depth maps, and log errors using cameras designed by our method with and without joint optimization, the RL method (DISeR), and off-the-shelf cameras, which are RealSense D450 and ZED2i. The depth and metrics are calculated in meters, and depth maps are capped at 1000 m. The off-the-shelf cameras fail with objects at long distances, whereas the cameras designed by our method and DISeR achieve desirable performance for all distance ranges.}
      \label{fig:result_stereo}
\end{figure*}
\begin{figure*}[t]
      \centering
      \includegraphics[width=\linewidth]{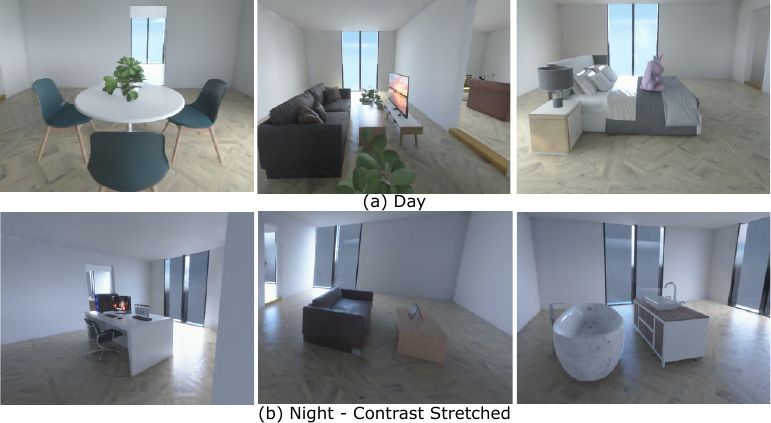}
      \caption{The example renders from our indoor virtual environment. The top 3 images (a) are captured in the daytime scenario (20 lux) using a lower camera gain (5 dB), and the lower 3 images (b) are captured in the nighttime scenario (2 lux) with a higher camera gain (15 dB). The nighttime images have stronger noise due to higher camera gain.}
      \label{fig:virtual_env}
\end{figure*}
\begin{figure}
      \centering
      \includegraphics[width=\linewidth]{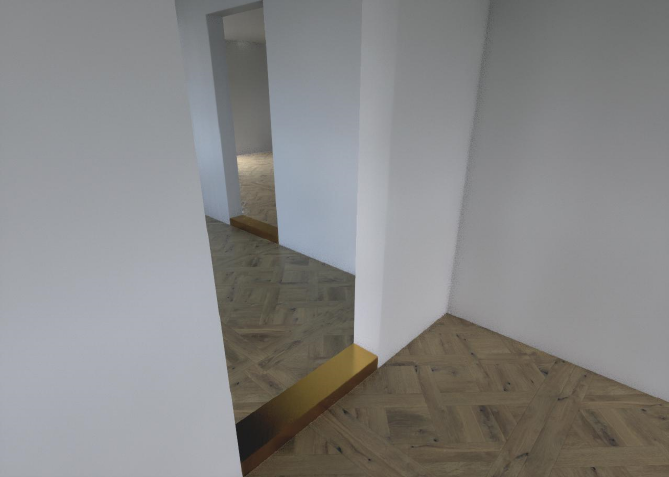}
      \caption{The low-height gold-colored thresholds act as obstacles that may pose a danger to the users.}
      \label{fig:obstacle}
\end{figure}
\begin{figure*}
      \centering
      \includegraphics[width=\linewidth]{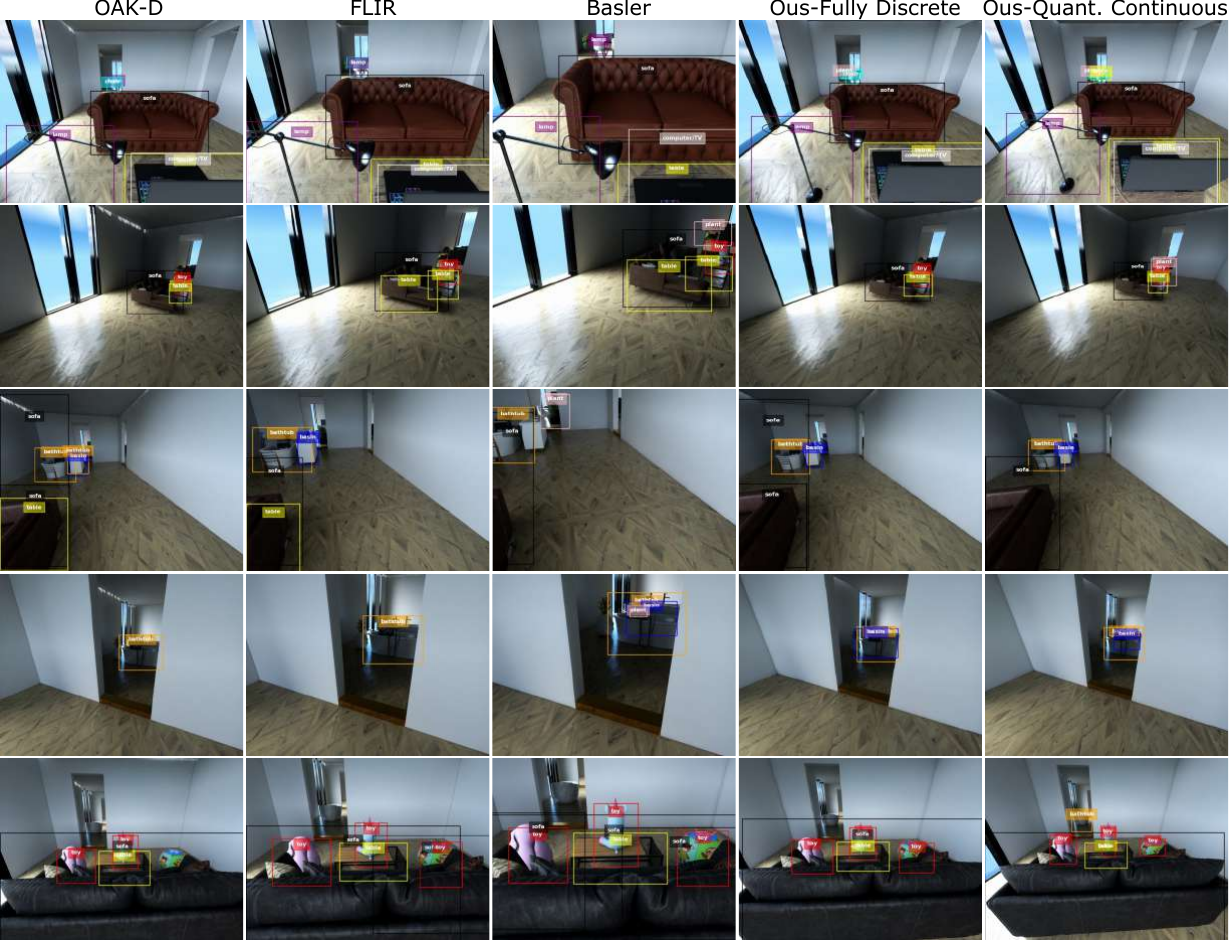}
      \caption{Comparison of object detection performance using cameras designed by our method and the off-the-shelf cameras. The cameras designed by our method show improved performance with small objects, objects at long distances, and objects that are partly occluded by optimizing the FOV and pixel size to obtain a more suitable effective resolution and signal-to-noise ratio for the task.}
      \label{fig:OD_results}
\end{figure*}
\begin{figure*}
      \centering
      \includegraphics[width=\linewidth]{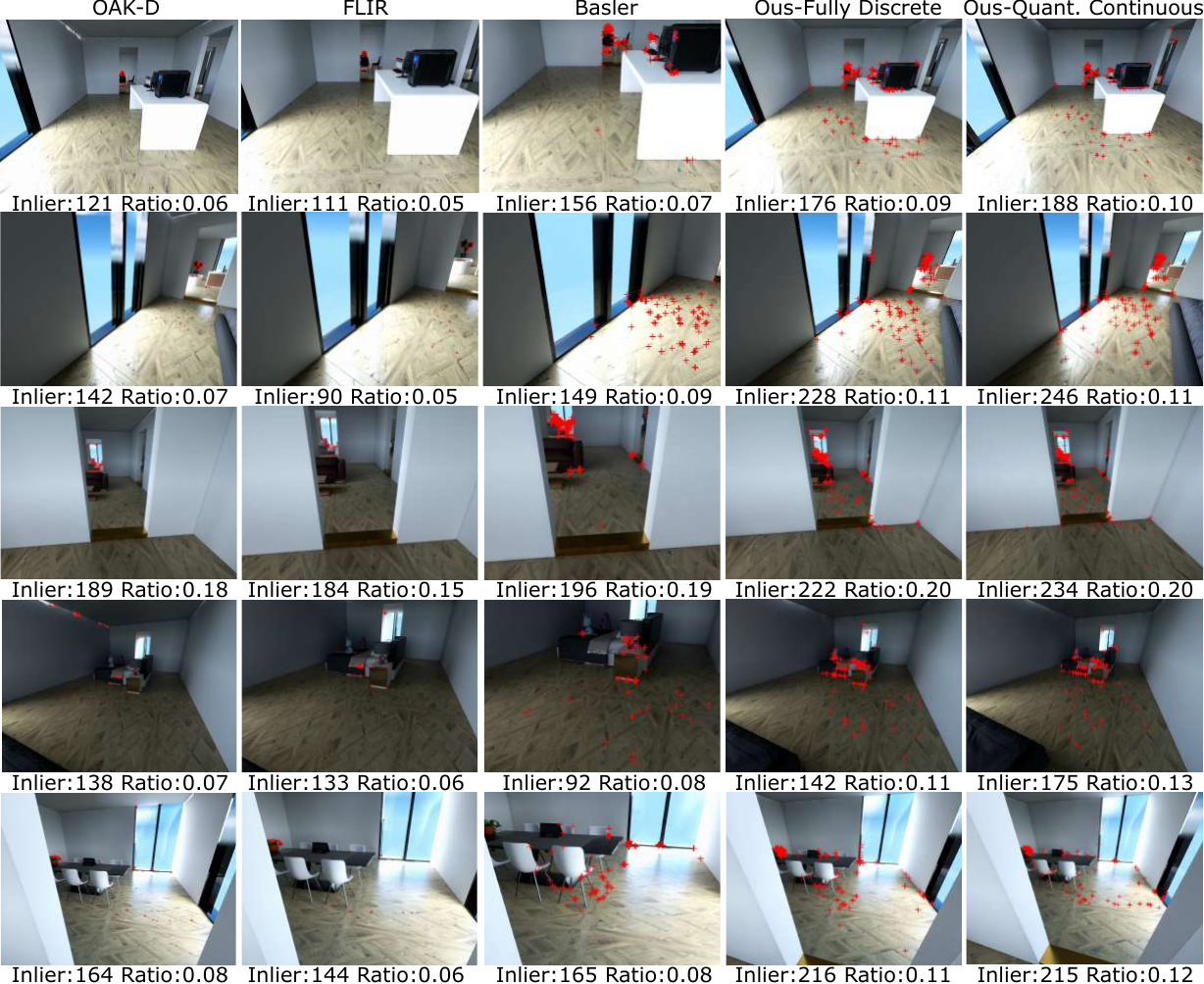}
      \caption{Comparison of feature extraction and matching with images captured by cameras designed by the proposed method and the off-the-shelf cameras. We display the features that are successfully matched with the features in the next frame and filtered (inliers) on the images, which shows that the cameras designed by our method contain the highest number of inliers.}
      \label{fig:FE_results}
\end{figure*}

\smallskip \noindent \textbf{Offspring Generation} The offspring generation process contains three steps: parent selection, crossover, and mutation. We select the top solutions from the current generation as parents for offspring generation. The number of parents is chosen as 50\% or 60\% of the size of the population, 50\% for an even number of solutions and 60\% for an odd number of solutions. For example, we use the top 3 solutions out of a total of 5 solutions per generation for the stereo camera design problem and use the top 5 solutions out of a total of 10 solutions per generation for the monocular design problem. We compare the optimization curves from 3 different portions of populations used for offspring generation (40\%, 60\%, 80\%) in the depth estimation task. The results are displayed in Fig.~\ref{fig:parent_num}, which indicate that although the final performance is not affected when using a large number of training steps, using 60\% of the population to generate offspring yields the fastest and smoothest convergence. Since using 40\% of the population as parents emphasizes exploitation over exploration, the optimizer may get stuck at local optima with fewer training steps. On the other hand, using 80\% of the population as parents creates more diverse offspring, which emphasizes exploration over exploitation and leads to slower convergence. Hence, we choose an intermediate percentage for offspring generation that is roughly half of the solution number per generation.

We then apply a uniform crossover scheme using the selected parents to produce the offspring for the next generation of solutions, indicating that each parameter for the offspring is randomly selected from the parents. For mutation, we apply a multiplication factor from the same range, which is a random number between 0.8 to 1.2, for all the parameters in our experiments and apply an addition value whose range is customised for different parameters depending on their available design space. All these hyperparameters used to implement the genetic algorithm are selected empirically.

\subsection{Parameter Initialization}
In our experiments, all the camera parameters that are optimized by the genetic algorithm are initialized with random values within their design space. However, we compare the optimization results using the proposed method with different initialization by setting the camera parameters in the depth estimation task to be initialized at their smallest values (baseline: 0.01 m, FOV: 50\textdegree) and their largest values in the design space (baseline: 3 m, FOV: 120\textdegree).

The results are displayed in Tab.~\ref{tab:initial}. The experiment uses the same hyperparameters for the genetic algorithm as described in Sec.~\ref{sec:hyperparam} to optimize the camera parameters. The perception network is jointly trained during optimization, indicated by \textcolor{green}{$\bigcdot$} in the table. The results indicate that the initialization of camera parameters does not have a significant impact on the final camera parameters and the downstream task performance.

In addition, we illustrate the training curves of the above-mentioned 3 camera parameter initialization schemes in Fig.~\ref{fig:ga_init}, which shows that random generation results in faster convergence compared to initialising from extreme values. This is because the optimal solution is usually not the extreme values, and starting from random values between the extremes gives values that are relatively closer to the optimal solution. It is observed that starting from the smallest parameter values results in the slowest convergence. This is because the offspring at every iteration is generated from the top solutions of the previous generation with a mutation process, where the parent solutions are multiplied by a random factor as the first mutation step. Therefore, smaller parameters change on a smaller scale compared to larger values, which slows down the exploration process and leads to slower convergence. 

\section{Additional Results on Stereo Camera Design}
We provide additional qualitative results for our stereo camera design experiment in Fig.~\ref{fig:result_stereo}. The figure displays the captured left images, the estimated depths, and the log error between the estimated maps and the ground-truth maps. We compare the images and results by using the stereo cameras designed by the proposed method with and without joint optimization, the Reinforcement Learning (RL) method (DISeR)~\cite{klinghoffer2023diser}, and two off-the-shelf cameras, which are Intel RealSense D450 and ZED 2i. The configurations of these cameras are listed in Tab.~1 of the main paper. The metrics displayed in this figure are the same as Tab.~1 of the main paper, which is log error and RMSE error in meters.

The results show that in our application scenario of the outdoor environment, the off-the-shelf cameras get low performance since their baselines are relatively low (0.095 m and 0.12 m), but many objects, such as the buildings and the footbridge, are far from the stereo camera. However, it is observed that these off-the-shelf cameras perform well in short distances, indicating that they can be beneficial for an application that does not involve long-distance objects. On the contrary, the cameras designed by our method and DISeR perform well across both long and short distances.

\section{Additional Details on Monocular Camera Design}
\subsection{Environment}
We construct the indoor virtual environment in UE 5 with a procedural generation technique, which generates random floorplans and object locations. The size of the environment is configured to be 15 m in width and length and 3 m in height for our experiment, and adjusting to different dimensions is trivial. However, each room in the environment needs to have a minimum length of 5 m to fit the furniture. The environment contains objects from 10 classes, which are sourced from the UE marketplace, encompassing 10 classes: sofa, bed, table, chair, bathtub, bathroom basins, computer/TV, plant, lamp, and toy.

We illustrate some example renders from our virtual environment in Fig.~\ref{fig:virtual_env}, including a comparison of the day and night design scenarios used to validate our method.

\subsection{Image Sensor Catalog}
The image sensor catalog we collected contains 43 image sensors, 28 of which are from Sony, 10 from Onsemi, 3 from Luxima, and 2 from CMOSIS. The pixel sizes of these sensors vary from 1.12 $\mu$m to 9 $\mu$m. The smallest sensor has a dimension of 3.07 mm$\times$2.3 mm, while the largest has a dimension of 16.13 mm$\times$12.04 mm.

\subsection{Obstacle Placement}
To restrict the camera's Field-of-View (FOV), we place low-height gold-colored thresholds at the entrance of all the rooms in our virtual environment. An example of the threshold is shown in Fig.~\ref{fig:obstacle}. The thresholds act as obstacles that may put the users at risk. They are configured to be interactive actors in UE, making the auto-agent aware of them even though they are not captured within the FOV of the scene capture camera.

\subsection{Qualitative Results}
We visualise the performance of object detection, as well as the feature extraction and matching task, with images captured by the cameras designed by the proposed method and the off-the-shelf machine/robotic cameras in Fig.~\ref{fig:OD_results} and Fig.~\ref{fig:FE_results} respectively. The off-the-shelf cameras are the same cameras used to validate our simulator in Sec.~\ref{sec:validation_task}, which are the OAK-D Pro Wide camera, FLIR Flea3 camera, and Basler Dart camera.

